%% file: root.tex
\documentclass[letterpaper, 10 pt, conference]{ieeeconf}  

\input{packages.tex}

\title{\LARGE \bf
A Convex Formulation of Frictional Contact for the Material Point Method and Rigid Bodies
}

\author{Zeshun Zong$^{1}$, Chenfanfu Jiang$^{1}$, Xuchen Han$^{2}$
\thanks{$^{1}$Zeshun Zong {\tt\small zeshunzong@math.ucla.edu} and Chenfanfu Jiang {\tt\small cffjiang@math.ucla.edu} are with the Department of Mathematics, University of California, Los Angeles, USA}%
\thanks{$^{2}$Xuchen Han {\tt\small xuchen.han@tri.global} is with Toyota Research Institute, USA}%
}

\begin{document}

\maketitle
\thispagestyle{empty}
\pagestyle{empty}

\input{abstract}


\input{introduction}
\input{previous_work}
\input{outline}
\input{mathematical_formulation}

\input{contact_point}
\input{constitutive_model}
\input{results}
\input{future_work}
\input{conclusion}





\section*{ACKNOWLEDGMENT}
We extend our gratitude for the support from the Dynamics \& Simulation team at Toyota Research Institute. We thank Damrong Guoy and Sean Curtis for visualization and geometry support and Alejandro Castro for helpful discussions.

\bibliographystyle{IEEEtran} 
\bibliography{root}

\end{document}

%% file: packages.tex
\usepackage{amsmath} 
\usepackage{amsfonts}
\usepackage{mathtools} 
\usepackage{amssymb}
\usepackage{graphicx}
\usepackage{bm}  
\usepackage[colorinlistoftodos, textsize=footnotesize]{todonotes}
\usepackage[colorlinks=true, allcolors=black]{hyperref}
\usepackage{setspace}
\usepackage{algorithm}      
\usepackage{algpseudocode}  
\usepackage{tikz}
\usepackage{verbatim}
\usepackage{xcolor}
\usepackage{subcaption}
\usepackage{todonotes}
\usepackage{makecell}

\usepackage[normalem]{ulem}

\usepackage[thinc]{esdiff}

\usepackage{mathrsfs}
\usepackage{booktabs}

\newif\ifcompile

\DeclareMathOperator*{\argmin}{arg\,min}

\newcommand{\vf}[1]{{\bm{#1}}}
\newcommand{\mf}[1]{{\mathbf{#1}}}

\newcommand{\vecx}{{\vf{x}}}
\newcommand{\vecX}{{\vf{X}}}
\newcommand{\vecv}{{\vf{v}}}
\newcommand{\matF}{{\mf{F}}}
\newcommand{\matC}{{\mf{C}}}
\newcommand{\matR}{{\mf{R}}}
\newcommand{\matS}{{\mf{S}}}
\newcommand{\matU}{{\mf{U}}}

\newcommand{\matI}{{\mf{I}}}
\newcommand{\dw}{{\nabla w}}

\DeclareMathOperator{\Tr}{Tr}
\DeclareMathOperator{\dev}{dev}



%% file: abstract.tex
\begin{abstract}

In this paper, we introduce a novel convex formulation that seamlessly integrates the Material Point Method (MPM) with articulated rigid body dynamics in frictional contact scenarios. We extend the linear corotational hyperelastic model into the realm of elastoplasticity and include an efficient return mapping algorithm. This approach is particularly effective for MPM simulations involving significant deformation and topology changes, while preserving the convexity of the optimization problem. Our method ensures global convergence, enabling the use of large simulation time steps without compromising robustness. We have validated our approach through rigorous testing and performance evaluations, highlighting its superior capabilities in managing complex simulations relevant to robotics. Compared to previous MPM-based robotic simulators, our method significantly improves the stability of contact resolution — a critical factor in robot manipulation tasks. We make our method available in the open-source robotics toolkit, Drake. 
The supplemental video is available \href{https://youtu.be/5jrQtF5D0DA}{\textit{here}}.

\end{abstract}

%% file: introduction.tex
\section{INTRODUCTION}

The advent of high-capacity models in artificial intelligence heralds a new era in large-scale learning, with significant implications for robotics and embodied AI. It underscores the importance of harnessing extensive and varied datasets to drive the evolution of embodied AI technologies \cite{bib:rtx}. In this context, simulation emerges as a pivotal tool, not only facilitating the provision of such expansive datasets \cite{bib:bousmalis2018using}, but also playing a critical role in the verification and evaluation of policies generated by these advanced models \cite{bib:choi2021use, bib:wilbert2024colosseum}. This necessitates the development of sophisticated robotic simulators capable of simulating a broad spectrum of phenomena with high fidelity and demonstrating robustness across varied input parameters. 

Within the mechanical engineering and computer graphics communities, the Material Point Method (MPM) \cite{bib:sulsky1994particle, jiang2016material} has seen a significant rise in popularity over recent decades due to its ability to simulate a wide range of physical objects and phenomena with high accuracy \cite{bib:de2020material}. Although there have been notable attempts to integrate MPM into robotics simulation \cite{bib:hu2019chainqueen, bib:huang2021plasticinelab, bib:gu2023maniskill2}, it has not yet seen widespread adoption.
This is partly because MPM, which is grounded in continuum mechanics and perceives the world as a deformable continuous medium, presents a fundamental contrast to the discrete assumptions underlying rigid body dynamics, the conventional approach for robot simulation. In this work, we aim to address this dichotomy with the first implicit integration between deformable bodies discretized with MPM and rigid bodies through tight frictional contact coupling. With this novel approach, we aim to facilitate robust interactions between robots, which are simulated under the premise of rigid body dynamics, and their environments, represented through MPM. Our goal with this integration is to enhance the versatility of simulation technologies, unlocking new possibilities for data generation and policy verification in a wider array of settings.

\begin{figure}
\centerline{\includegraphics[width=1.0\columnwidth]{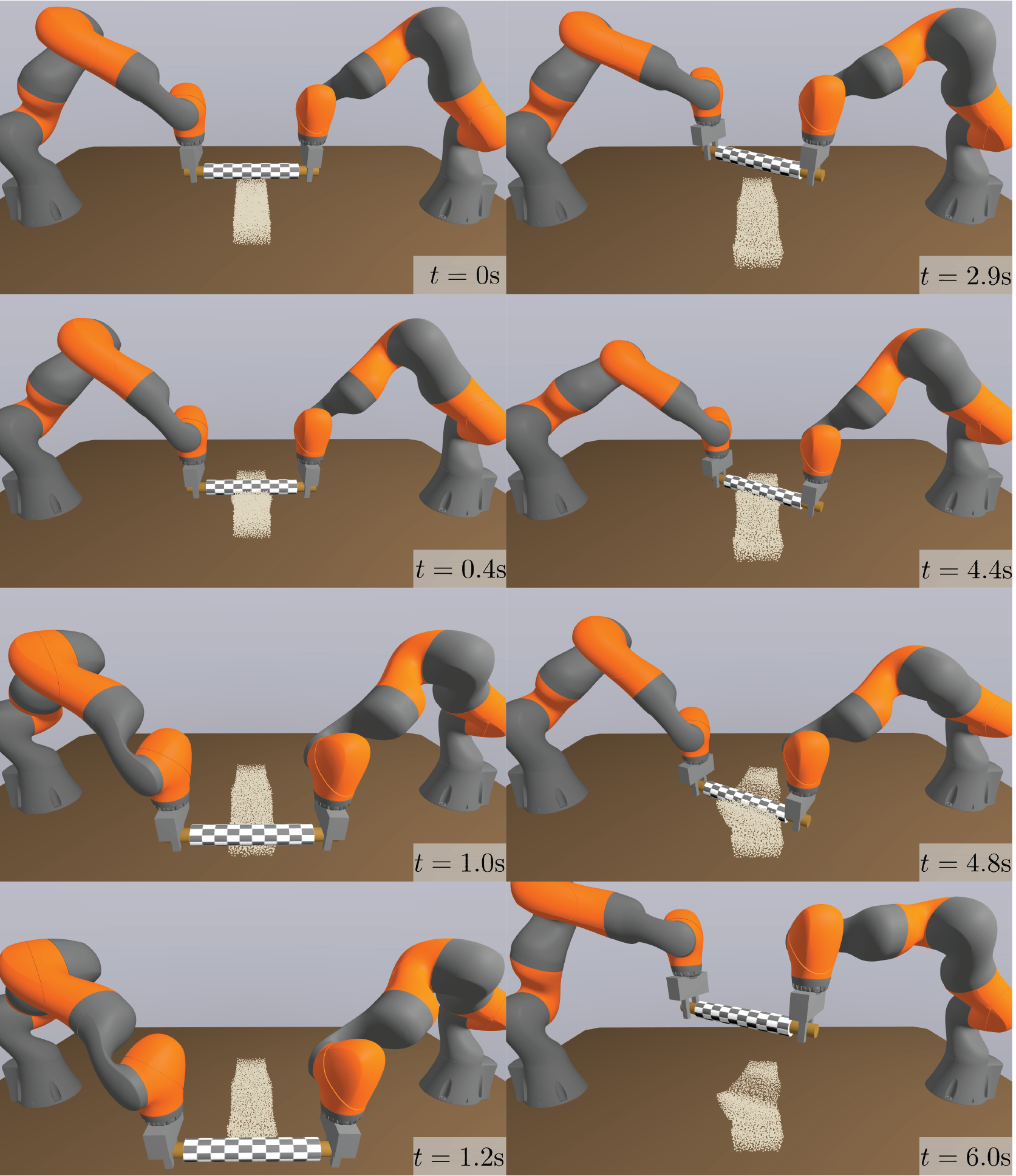}}
\caption{Rolling out dough with a rolling pin  (see supplemental video). Our two-way coupled solver captures the dough's deformation as well as the rolling pin's rotation driven by frictional contact with the dough.}
\label{fig:dough-rolling}
\end{figure}

%% file: previous_work.tex
\section{PREVIOUS WORK}
\subsection{Material Point Method}
MPM is a hybrid simulation method that combines Lagrangian particles carrying physical states and a background Eulerian grid for discretization of continuous fields. Originally developed as an extension to the particle-in-cell (PIC) methods for solid mechanics problems \cite{bib:sulsky1994particle}, MPM has been applied to many engineering problems such as the simulation of landslide \cite{bib:andersen2010modelling}, terramechanics \cite{bib:agarwal2019modeling}, and avalanche \cite{bib:trottet2022transition}. The adoption and development of MPM have significantly accelerated following its introduction to the computer graphics community \cite{bib:stomakhin2013material}, leading to its application in simulating a variety of phenomena, especially those characterized by elastoplastic behaviors. By employing specialized elastoplasticity constitutive models, MPM can simulate the dynamics of non-Newtonian fluids and foams \cite{bib:ram2015material, bib:yue2015continuum, 
bib:su2021unified}, porous media \cite{bib:tampubolon2017multi}, the phase change process \cite{bib:ding2019thermomechanical, bib:stomakhin2014augmented}, etc. Elastoplastic models have also been designed to simulate frictional contact among granular material \cite{bib:daviet2016semi, bib:klar2016drucker}, thin materials like clothing and hair \cite{bib:fei2018multi, bib:guo2018material, bib:jiang2017anisotropic}, and volumetric objects \cite{
bib:han2019hybrid}. Recent endeavors have sought to introduce MPM to robotics simulation \cite{bib:hu2019chainqueen, bib:huang2021plasticinelab, bib:gu2023maniskill2, bib:chen2023tacchi}, and have demonstrated its potential in robotics applications \cite{bib:shi2023robocook, bib:tuomainen2022manipulation}. However, robust treatment of frictional contact with rigid bodies with guaranteed stability is still lacking. Concurrently, the introduction of variational methods has aimed to enable larger time steps in MPM simulations while ensuring stability \cite{bib:gast2015optimization, bib:wang2020hierarchical}.
Our method follows a similar approach by adopting a variational framework but differs from previous work by formulating a convex optimization problem. The convexity of the problem guarantees global convergence and stability even in highly challenging scenarios.

\subsection{Frictional Contact}
In robotics simulation, especially for manipulation tasks, the accurate and stable resolution of frictional contact is crucial. However, rigid contact combined with Coulomb model of friction can lead to configurations, known as Painlev\'{e} paradoxes, where the solution at the force and acceleration level does not exist. A common mitigation strategy involves reformulating the problem at the velocity level into a non-linear complementarity problem (NCP). Many robotics simulators adopt the projected Gauss-Seidel (PGS) solvers to transfer collision impulses between objects in collision to address the NCP \cite{bib:bullet, bib:macklin2019small}. Despite its simplicity, PGS is theoretically known to converge exponentially slowly in the worst case \cite{bib:erleben2007velocity}, and, in practice, often fails to solve the NCP with the requisite accuracy, leading to instabilities \cite{bib:ferguson2021rigidipc}. This shortcoming becomes particularly critical in robot manipulation tasks, where precise object interaction is crucial for task success \cite{bib:kim2022ipc}. Efforts to integrate rigid bodies and MPM through impulse-based methods have been also explored across the engineering \cite{bib:zhao2023coupled}, computer graphics \cite{bib:han2019hybrid, bib:hu2018moving}, and robotics \cite{bib:huang2021plasticinelab, bib:gu2023maniskill2} communities, yet accuracy and stability challenges still persist in contact intensive scenarios.

In recent years, a wave of variational methods has significantly improved the state of the art of frictional contact resolution. Incremental Potential Contact (IPC) \cite{bib:li2020ipc} addresses nonlinear elastodynamics problems, providing non-penetration guarantees and has been extended to accommodate rigid bodies \cite{bib:ferguson2021rigidipc}, near-rigid bodies \cite{bib:lan2022affine}, and codimensional objects \cite{bib:li2021cipc}. Similar guarantees are offered by \cite{bib:howell2022dojo}, which employs an interior point method solver. Adopting a compliant contact approach, \cite{bib:castro2022unconstrained} builds on top of \cite{bib:anitescu2002} and 
\cite{bib:todorov2012mujoco} to formulate an unconstrained convex optimization problem for frictional contact and proposes the Semi-Analytical Primal (SAP) solver that guarantees global convergence. This method is extended by \cite{bib:han2023convex} to support deformable bodies modeled with Finite Element Method (FEM). In this work, we propose a further extension to incorporate MPM, enabling the robust simulation of a broader spectrum of materials and further enriching the domain of robotics simulation.

%% file: outline.tex
\section{OUTLINE AND NOVEL CONTRIBUTIONS}

To the knowledge of the authors, this work presents the first convex formulation for MPM that implicitly couples with articulated rigid bodies through frictional contact. Section \ref{sec:mathematical_formulation} describes the mathematical formulation of our framework and our discretization strategy. Section \ref{sec:mpm-contact} details how contact points are generated. In Section \ref{sec:constitutive_model}, We propose a novel elastoplastic model with a simple return mapping projection scheme that makes the convex problem quadratic and efficient to solve. Validation and comparison against state-of-the-art alternatives are presented in Section \ref{sec:results}. We make the implementation of our method publicly available for the robotics community as part of the open-source robotics toolkit Drake \cite{bib:drake}.

%% file: mathematical_formulation.tex
\section{Mathematical Formulation}\label{sec:mathematical_formulation}
The state of our system consists of generalized positions $\mf{q} \in \mathbb{R}^{n_q}$ and velocities $\mf{v} \in \mathbb{R}^{n_v},$ where $n_q$ and $n_v$ are the total number of generalized positions and velocities, respectively. A kinematic map $\mf{N}(\mf{q}) \in \mathbb{R}^{n_q \times n_v}$ relates the generalized positions and velocities by $\dot{\mf{q}} = \mf{N}(\mf{q}) \mf{v}.$ 
At a given configuration $\mf{q}$, we compute the set $\mathcal{C}(\mf{q})$ of contact points between pairs of bodies. Each contact point $j \in \mathcal{C}({\mf{q}})$ is characterized by its location $\vf{p}_j \in \mathbb{R}^3$, the contact normal $\hat{\vf{n}}_j\in \mathbb{R}^3$, the penetration distance $\phi_j>0$, and the contact velocity $\vf{v}_{c,j}\in \mathbb{R}^3$ expressed in a contact frame $C_j$ for which we arbitrarily choose the $z$-axis to coincide with the contact normal $\hat{\vf{n}}_j$. The contact velocity has a positive normal component when objects are moving toward each other, and is related to the generalized velocity via $\vf{v}_{c,j} = \mf{J}_j \mf{v}$ with $\mf{J}_j \in \mathbb{R}^{3\times n_v}$ being the Jacobian matrix. Collectively, we can stack all $n_c$ contact velocities and write $\mf{v}_c = \mf{J} \mf{v},$  where $\mf{J} \in \mathbb{R}^{3n_c\times n_v}$ is formed by stacking each corresponding Jacobian matrix $\mf{J}_j$. We describe how these contact quantities are computed in Section \ref{sec:mpm-contact}.

\subsection{Two Stage Implicit Time Stepping}
We discretize time into intervals of size $\Delta t$ to advance the dynamics of the system from $t_n$ to the next time step $t_{n+1} := t_n + \Delta t$ subject to frictional contact constraint, as 
\begin{align}\label{eq:momentum}
    \mf{M}(\mf{q}^{n}) (\mf{v}^{n+1} - \mf{v}^n) = &\Delta t \left(\mf{c}(\mf{q}^n, \mf{v}^n) + \mf{k}(\mf{q}^{n+1}, \mf{v}^{n+1})\right) \notag \\
    &+ \mf{J}(\mf{q}^n)^T \bm{\gamma}(\mf{v}^{n+1}),
\end{align}
\begin{equation}\label{eq:contact constraint}
    \begin{split}
        \mathcal{F}_j \ni \vf{\gamma}_j \perp \vf{v}_{c,j} - \hat{\vf{v}}_{c,j}(\phi_j(\mf{q}^n), \hat{\vf{n}}_j(\mf{q}^n)) \in \mathcal{F}_j^* \\
        \text{ where } j = 1, 2, ..., n_c,
    \end{split}
\end{equation}
\begin{align}
    &\mf{q}^{n+1} = \mf{q}^n + \Delta t \mf{N}(\mf{q}^n) \mf{v}^{n+1}. \label{eq:position-update}
\end{align}
$\mf{M} \in \mathbb{R}^{n_v \times n_v}$ is the mass matrix; $\mf{c}\in \mathbb{R}^{n_v}$ is Coriolis and gyroscopic forces; $\mf{k}\in \mathbb{R}^{n_v}$ includes all other non-contact generalized forces; $\bm{\gamma} \in \mathbb{R}^{3n_c}$ stacks contact impulses $\vf{\gamma}_j \in \mathbb{R}^{3}$ at each contact point; $\hat{\vf{v}}_{c,j}$ is the stabilization velocity of the $j$-th contact point \cite{bib:castro2022unconstrained}, which is a function of the penetration distance $\phi_j$ and the contact normal $\hat{\vf{n}}_j$ at configuration $\mf{q}^n$; $\mathcal{F}_j$ is the friction cone of the $j$-th contact point and $\mathcal{F}_j^*$ is its dual. Note that both the contact impulse $\bm{\gamma}$ and internal elastic forces of deformable bodies (included in $\mf{k}$) are treated implicitly, which is crucial for stability. 

Similar to \cite{bib:castro2022unconstrained} and \cite{bib:han2023convex}, we adopt a two-stage scheme where we first solve
for the free motion velocities $\mf{v^*}$ the system would have in the
absence of contact, according to
\begin{equation}\label{eq:v_star}
\mf{m}(\mf{v}^*) = \mf{0},
\end{equation}
where we define the momentum residual $\mf{m}(\mf{v})$ to be
\begin{equation}\label{eq:momentum_residual}
\mf{m}(\mf{v})=\mf{M}(\mf{q}^n)(\mf{v} - \mf{v}^n) - \Delta t \left(\mf{c}(\mf{q}^n, \mf{v}^n) + \mf{k}(\mf{q}(\mf{v}), \mf{v})\right).
\end{equation}
In the second stage, we account for the effect of frictional contact by solving for $\mf{v}^{n+1}$ in the linearized balance of momentum around $\mf{v}^*$
\begin{equation}\label{eq:linearized_momentum_balance}
    \mf{A}(\mf{v}^{n+1} - \mf{v}^*) = \mf{J}(\mf{q}^n) \bm\gamma(\mf{v}^{n+1}) 
\end{equation}
subject to the same friction cone constraints (2), where $\mf{A} = \partial \mf{m} / \partial \mf{v} |_{\mf{v}^*}$.
Following \cite{bib:castro2022unconstrained}, \eqref{eq:linearized_momentum_balance} can be cast as an unconstrained optimization problem
\begin{equation}\label{eq:optimization_problem}
    \min_{\mf{v}} \ell(\mf{v}) = \frac{1}{2}\Vert\mf{v}-\mf{v}^*\Vert_{\mf{A}}^2 +
	\ell_c(\mf{v}_c(\mf{v})),
\end{equation}
where $\Vert\mf{x}\Vert_{A}^2 = \mf{x}^T\mf{A}\,\mf{x}$ and $\ell_c$ is
a convex contact potential. 
Notice that since $\ell_c$ is convex, $\ell$ is strictly convex and has a unique global minimum if $\mf{A}$ is symmetric positive-definite (SPD). We refer to \cite{bib:castro2022unconstrained} and \cite{bib:castro2023theory} for details about the convex contact potential. We explain how to compute $\mf{v}^*$ and $\mf{A}$ and show that $\mf{A}$ is SPD in the next section. 

\subsection{Deformable Body Discretization with MPM}\label{sec:mpm}
We represent all deformable bodies interacting with rigid bodies as an elastoplastic continuum, whose kinematics is represented by the deformation map $\vf{x} = \vf{\phi}(\vf{X}, t)$, mapping reference positions $\vecX$ in the reference domain $\Omega \subset \mathbb{R}^3$ to their current positions $\vecx$ at time $t$. The deformation gradient $\matF = \partial \vf{\phi} / \partial \vf{X}$ measures the amount of local deformation of the material and can be multiplicatively decomposed into elastic and plastic parts as $\matF = \matF^E \matF^P$ . The total elastic energy of the continuum is 
\begin{equation}
    E_{\text{elastic}} = \int_\Omega \Psi(\matF^E(\vecX)) d\vecX. \label{eq:total-elastic-e}
\end{equation}
Here $\Psi$ is the elastic energy density function of the chosen constitutive model. As typical with spatial discretization using MPM, we use a background Cartesian grid with grid spacing $h$ as the computational mesh and track material states on particles. In particular, at each time step $t_n$, we track particle position $\vf{x}^n_p$, velocity $\vf{v}^n_p$, total deformation gradient $\mf{F}^n_p$, plastic deformation gradient ${\mf{F}_p^P}^n$, mass $m_p$, initial volume $V^0_p$, and affine velocity $\mf{C}^n_p$ \cite{bib:jiang2015affine}. Let $w_{ip}^n$ and $\dw^n_{ip}$ denote the interpolation weight between particle $p$ and grid node $i$ and its gradient at $t_n$.
Mass and velocity are transferred to the $i$-th grid node with position $\vf{x}_i$ using 
\begin{equation} \label{eq:p2g}
\begin{split}
    m_i &= \sum_p w_{ip}^n m_p,\\
    \vecv_i^n &= \frac{1}{m_i}\sum_p w_{i p}^n m_p\left(\vecv_p^n+\matC_p^n\left(\vecx_i-\vecx_p^n\right)\right),
    \end{split}
\end{equation}
At each time step, we treat the positions and velocities of all grid nodes with nonzero mass, in additional to rigid degrees of freedom (DoFs), as generalized position and velocity in \eqref{eq:momentum}. Formally, we write 
\begin{equation} \label{eq:generalized_coordinates}
\mf{q} = [\mf{q}^T_\text{rigid}, \mf{q}^T_\text{MPM}]^T, \quad
    \mf{v} = [\mf{v}^T_\text{rigid}, \mf{v}^T_\text{MPM}]^T. 
\end{equation}
Consequently, the kinematic map $\mf{N}$ is the identity map $\mf{I}_{3n_i \times 3n_i}$ when restricted to the MPM DoFs, where $n_i$ is the number of grid nodes with nonzero mass at $t_n$.

With this choice of generalized DoFs, we observe that \eqref{eq:momentum_residual}, in the absence of contact, decouples into equations for rigid body DoFs and MPM DoFs. We refer to \cite{bib:featherstone2008_rigid_body_dynamics_algorithms} and \cite{bib:castro2022unconstrained} for the rigid body free motion velocity computation and focus on the equations for the MPM DoFs. As standard with MPM, we apply mass lumping and define the mass matrix $\mf{M}_{\text{MPM}}$ as the diagonal matrix with $m_i \mf{I}_{3\times3}$ as the $i$-th diagonal block. Hence, \eqref{eq:v_star} further decouples into
\begin{equation}
    m_i^n (\vecv^*_i - \vecv^n_i) - \Delta t ( m_i^n \vf{g}+ \vf{k}_i(\mf{v}_{\text{MPM}}^*) ) = \vf{0},\label{eq:grid-node-momentum}
\end{equation}
where $\vf{g}$ is the gravity vector. The elastic internal force acting grid $i$, $\vf{k}_i$, is computed as
\begin{align}
    \vf{k}_i(\mf{v}^*_{\text{MPM}}) &= - \frac{\partial E_{\text{elastic}}}{ \partial \vecx_i}(\mf{v}^*_{\text{MPM}}) \\
    &= - \frac{1}{\Delta t} \sum_p V^0_p \frac{\partial \Psi(\matF^E_p(\mf{v}^*_{\text{MPM}}))}{\partial \vecv_i},
\end{align}
where the total elastic energy in \eqref{eq:total-elastic-e} is discretely approximated as $\sum_p V^0_p \Psi(\matF^E_p)$.
The set of equations is closed with the following deformation gradient update rule
\begin{equation}\label{eq:F_update}
    \begin{split}
        \mf{F}^E_p(\mf{v}_{\text{MPM}}) &= \mf{F}_p(\mf{v}_{\text{MPM}}) \left({{\mf{F}_p^P}^n}\right)^{-1}, \\
 \matF_p (\mf{v}_{\text{MPM}})&=\left(\mf{I}+\Delta t \sum_i \vecv_i\left(\nabla w_{i p}^n\right)^T\right) \matF_p^n.
    \end{split}
\end{equation}
Further, \eqref{eq:grid-node-momentum} can be cast as an optimization problem
\begin{align}\label{eq:mpm_optimization}
    \mf{v}_{\text{MPM}}^* 
    =\argmin_{\mf{v}} E_{\text{MPM}}(\mf{v}),
\end{align}
where we define 
\begin{equation}
\label{eqn:mpm-objective-function}
    \begin{split}
        E_{\text{MPM}}(\mf{v}) &= \frac{1}{2}\|\mf{v} - \mf{v}_{\text{MPM}}^n\|^2_{\mf{M}}  \\&- \Delta t \mf{v}^T \mf{M}_{\text{MPM}} \mf{g} + \sum_p V^0_p \Psi(\matF^E_p(\mf{v}))
    \end{split}
\end{equation}
with $\mf{g}$ being the stacked gravity vector.
Since \eqref{eq:F_update} is linear in $\mf{v}_{\text{MPM}}$ and $\mf{M}_{\text{MPM}}$ is SPD, if $\Psi$ is convex, the problem is convex and can be robustly solved with Newton's method with line search. Furthermore, $\nabla^2 E_{\text{MPM}}(\mf{v}^*_\text{MPM})$ is the $\mf{A}$ matrix sought in \eqref{eq:linearized_momentum_balance} for the MPM DoFs and is guaranteed to be SPD. In particular, when $\Psi$ is quadratic in $\mf{F}$ (see Section \ref{sec:constitutive_model}), the objective is quadratic and the optimization problem can be solved by solving a single linear system of equations. Equipped with $\mf{v}^*$ and $\mf{A}$, we solve the optimization problem \eqref{eq:optimization_problem} for $\mf{v}^{n+1}$ with the SAP solver from \cite{bib:castro2022unconstrained} and extract grid velocities $\vf{v}_i^{n+1}$. Particle positions and velocities are then updated according to
\begin{equation}\label{eq:particle_update}
    \vf{v}_p^{n+1} = \sum_i w_{ip}^n \vf{v}_i^{n+1}, \quad
    \vf{x}_p^{n+1} = \vf{x}_p^{n} + \Delta t \vf{v}_p^{n+1}.
\end{equation}
We then update $\mf{F}_p^{n+1}$ using \eqref{eq:F_update} and $\mf{C}_p^{n+1}$ following \cite{bib:jiang2015affine}. We describe the update rule for ${\mf{F}_p^{P}}^{n+1}$ in Section \ref{sec:constitutive_model}. We refer to \cite{jiang2016material} for additional MPM details.

%% file: contact_point.tex
\section{Contact Point Computation} \label{sec:mpm-contact}
We use the pressure field contact model from \cite{bib:masterjohn2022velocity} to sample contact points between rigid geometries in contact. Deformable bodies, by virtue of the hybrid Eulerian-Lagrangian nature of MPM, inherently handles self-collision \cite{bib:jiang2017anisotropic, bib:han2019hybrid}. Therefore, we do not sample contact points between deformable bodies. In this section, we detail the process of generating contact points between rigid geometries and deformable bodies discretized with MPM.

At each time step $t_n$, we register a contact point for each MPM particle $p$ overlapping with a rigid geometry $b$. The contact point position $\vf{p}_p$ is defined as the particle position $\vf{x}_p^n$, and the contact normal is the unit vector pointing from $\vf{q}_p$ to $\vecx_p^n$, where $\vf{q}_p$ is the nearest point of $\vecx_p^n$ on the surface of $b$ computed using sign distance field for analytical shapes and point to mesh distance query for meshes. The penetration distance is computed as $\|\vecx^n_p-\vf{q}_p\|$. Using \eqref{eq:generalized_coordinates}, the contact velocity of the $j$-th contact point can be written as
\begin{equation}
    \vf{v}_{c, j} = \mf{J}_{j, \text{rigid}} \mf{v}_ \text{rigid} + \mf{J}_{j, \text{MPM}} \mf{v}_ \text{MPM}.
\end{equation}
$\mf{J}_{j, \text{rigid}}$ is computed by compositing the contribution of each node in the kinematic path of the articulated rigid body up to the body in contact \cite{bib:featherstone2008_rigid_body_dynamics_algorithms}. Using \eqref{eq:particle_update}, we compute $\mf{J}_{j, \text{MPM}}$ as the $3$-by-$3n_i$ block matrix with nonzero $3$-by-$3$ blocks with value $-w_{ip}\mf{I}_{3\times3}$ for block column $i$ corresponding to grid $i$. Because of the local support of the interpolation function, $w_{ip}$ is only nonzero for grid nodes $i$ in the influence of particle $p$. Therefore, the contact Jacobian $\mf{J}$ is in general sparse and we leverage this to drastically reduce the size of problem \eqref{eq:optimization_problem} by adopting the Schur complement strategy detailed in \cite{bib:han2023convex}.

%% file: constitutive_model.tex
\section{Corotational Model with Plasticity} \label{sec:constitutive_model}
We propose a new elastoplastic material model that satisfies the convexity requirement described in Section \ref{sec:mpm} by making careful approximation to established elastic and plastic models from the engineering literature. With Lam\'{e} parameters $\mu$ and $\lambda$, the energy density of our model follows that from \cite{bib:han2023convex}
\begin{equation}\label{eq:energy_density}
    \Psi(\matF^E) = \mu || \hat{\mf{E}}||_F^2 + \frac{\lambda}{2} \Tr(\hat{\mf{E}})^2,
\end{equation}
where $\hat{\mf{E}} = \frac{1}{2} (\matR_0^T \matF^E + {\matF^E}^T \matR_0) - \matI$ is the linearized corotational strain and $\matR_0$ is the rotation matrix from the polar decomposition of ${\mf{F}^E}^n$, the elastic deformation gradient from the previous time step. As noted by \cite{bib:han2023convex}, this energy density is an $\mathcal{O}(\Delta t)$ approximation to the corotational model and is a good approximation to the St.Venant-Kirchhoff model when the principal stretches of $\mf{F}^E$ are small. Furthermore, the energy density is quadratic in $\matF^E$, making problem \eqref{eq:mpm_optimization} quadratic and efficient to solve. 

To fully exploit MPM's potential in simulating extreme deformations and topological changes, we design a plasticity model for this energy density. The Kirchhoff stress of the model is given by 
\begin{align}
    \bm{\tau} &= \frac{d\Psi}{d\matF^E}{\mf{F}^E}^T = \mf{PVR}^T, \\
     \mf{P} &=  \frac{d\Psi}{d\matF^E}= \matR_0 (2\mu \mf{\hat{E}} + \lambda \Tr(\mf{\hat{E}}) \matI),
\end{align}
where $\mf{F}^E = \mf{RV}$ is the polar decomposition of the elastic deformation gradient. We approximate von Mises yield criterion
$\|\dev(\bm{\tau})\|_F \leq \eta$
where $\dev(\bm{\tau}) = \bm{\tau} - \frac{1}{3} \Tr(\bm{\tau}) \mf{I}$ is the deviatoric part of the Kirchhoff stress and $\eta$ is the yield stress.
This approximation replaces the Kirchhoff stress $\bm{\tau}$ with $\mf{P}\mf{R}_0$, omitting the $\mf{V}$ term by again assuming principal stretches of $\mf{F}^E$ are small and approximating $\mf{R}$ with $\mf{R}_0$. Consequently, the adopted yield criterion is
\begin{equation}
    ||\dev(\mf{P}\matR_0^T)||_F \leq \eta.\label{eq:yield-criterion}
\end{equation}
One can show that this is equivalent to 
\begin{equation}
    ||\dev(\matS)||_F \leq \frac{\eta}{2\mu},
    \label{eq:yield-criterion2}
\end{equation}
with $\matS$ defined as $\hat{\mf{E}} + \mf{I}$, which 
can be further simplified as 
\begin{align}
    \sum_{i=1,2,3} \left(\sigma_i - \frac{\sigma_1 + \sigma_2 + \sigma_3}{3}\right)^2  \leq \left( \frac{\eta}{2\mu} \right) ^2,
\end{align}
where $\bm{\sigma} = [\sigma_1, \sigma_2, \sigma_3]^T$ are the eigenvalues of the symmetric matrix $\matS$.
This describes the interior of the infinite cylinder with axis parallel to $[1,1,1]^T$ and radius $\frac{\eta}{2\mu}.$ Following the von Mises model, we orthogonally project eigenvalues of $\mf{S}$ outside the cylinder onto the surface of the cylinder. That is, the projected eigenvalues $\vf{\tilde{\sigma}} = [\tilde{\sigma}_1, \tilde{\sigma}_2,\tilde{\sigma}_3]^T$ satisfy
\begin{equation}
    \vf{\tilde{\sigma}} =\vf{\bar{\sigma}} +  \min\left\{\frac{\eta/2\mu}{||\vf{\sigma} - \vf{\bar{\sigma}}|| },1.0\right\} \cdot (\vf{\sigma} -\vf{\bar{\sigma}})\label{eq:return_mapping}
\end{equation}
where $\vf{\bar{\sigma}} = [\bar{\sigma}, \bar{\sigma}, \bar{\sigma}]^T$ and ${\bar{\sigma}} = (\sigma_1 + \sigma_2 + \sigma_3) / 3$. 
We then build $\mf{\tilde{S}} = \matU \operatorname{diag}( \vf{\tilde{\sigma}}) \matU^T$ using the eigenvectors $\mf{U}$ of $\matS$, and finally we compute the projected elastic deformation gradient $\tilde{\matF}^E = \matR \tilde{\mf{V}}$ by solving for $\tilde{\mf{V}}$ in the $6\times6$ linear system
\begin{equation}
    \tilde{\matS} = \frac{1}{2} (\matR_0^T \matR \tilde{\mf{V}} +  \tilde{\mf{V}} \matR^T \matR_0).
    \label{eq:S-tilde=RF+FR}
\end{equation}
This plastic projection is lagged and only performed for each particle at the end of the time step to compute ${\mf{F}_p^P}^{n+1}$ after $\mf{F}_p^{n+1}$ is computed using \eqref{eq:F_update}. Problem \eqref{eq:mpm_optimization} is solved using the elastic deformation gradient in \eqref{eq:F_update} without plastic flow and remains quadratic. We summarize the return mapping procedure to compute ${\mf{F}_p^P}^{n+1}$ in Algorithm \ref{alg:returnmap}. The particle subscript $p$ is omitted for simplicity.

\algrenewcommand\algorithmicrequire{\textbf{Input:}}
\algrenewcommand\algorithmicensure{\textbf{Output:}}
\begin{algorithm}
\caption{Plastic deformation gradient update}\label{alg:returnmap}
\begin{algorithmic}[1]
 \Require{$\mf{F}^{n+1}$, ${\mf{F}^P}^n$}
 \Ensure{${\mf{F}^P}^{n+1}$}
\State Compute the trial elastic deformation gradient $\matF^E = \mf{F}^{n+1} \left({{\mf{F}^P}^n}\right)^{-1}$ and its polar decomposition $\matF^E = \mf{RV}$.
\State Compute $\mf{S}$ from $\mf{F}^E$ and compute its eigenvalues $\bm{\sigma}$.
\State Project $\bm{\sigma}$ to obtain $\bm{\tilde{\sigma}}$ using \eqref{eq:return_mapping}.
\State Build $\tilde{\matS}$ and solve for $\tilde{\matF}^E$ using \eqref{eq:S-tilde=RF+FR}. 
\State ${\mf{F}^P}^{n+1} \gets \left(\tilde{\matF}^E \right)^{-1} \mf{F}^{n+1}$.
\end{algorithmic}
\end{algorithm}

We now have all the ingredients required to summarize the entire time-stepping strategy of our method in Algorithm \ref{alg:pipeline}, where we denote the state of all MPM particles at time $t_n$ as $\mathcal{P}^n = \{(\vecx_p^n, \vecv_p^n, \matF_p^n, \mf{C}_p^n) \text{ for each particle }p\}$.
\begin{algorithm}[H]    
    \caption{Temporal advancement from $t_n$ to $t_{n+1}$}\label{alg:pipeline}
    \begin{algorithmic}[1]
        \Require{$\mf{q}^n_\text{rigid}, \mf{v}^n_\text{rigid}, \mathcal{P}^n$}
        \Ensure{ $\mf{q}^{n+1}_\text{rigid}, \mf{v}^{n+1}_\text{rigid}, \mathcal{P}^{n+1}$ }
        \State Transfer particle state to grid state using \eqref{eq:p2g}.
        \State Form $\mf{q}^n$ and $\mf{v}^n$ in \eqref{eq:momentum_residual} using \eqref{eq:generalized_coordinates}.
        \State Compute contact constraints in \eqref{eq:contact constraint} and compute contact Jacobian $
        \mf{J}(\mf{q}^n)$ in \eqref{eq:linearized_momentum_balance}, following Section \ref{sec:mpm-contact}.
        \State Solve the equations corresponding to the rigid DoFs in \eqref{eq:v_star} as in \cite{bib:castro2022unconstrained} to obtain $\mf{v}^*_\text{rigid}$ and $\mf{A}_\text{rigid}.$
        \State Solve problem \eqref{eq:mpm_optimization} with Newton's method to obtain $\mf{v}^*_\text{MPM}$ and $\mf{A}_\text{MPM}$ as the Hessian of the objective $E_\text{MPM}$.
        \State Form $\mf{v}^* = [\mf{v}^{*}_{\text{rigid}}, \mf{v}^{*}_{\text{MPM}}]^T$ and $\mf{A} = \operatorname{diag}(\mf{A}_\text{rigid}, \mf{A}_\text{MPM})$ in \eqref{eq:linearized_momentum_balance}.
        \State Solve for $ [\mf{v}^{n+1}_{\text{rigid}}, \mf{v}^{n+1}_{\text{MPM}}]^T = \mf{v}^{n+1}$ in \eqref{eq:linearized_momentum_balance} as in \cite{bib:castro2022unconstrained}.
        \State  $\mf{q}^{n+1}_\text{rigid} \leftarrow \mf{q}^{n}_\text{rigid} + \Delta t \mf{N}(\mf{q}^{n}_\text{rigid})\mf{v}^{n+1}_{\text{rigid}}.$
        \State Update $\mathcal{P}^{n+1}$ using \eqref{eq:particle_update} and Algorithm \ref{alg:returnmap}.
    \end{algorithmic}
\end{algorithm}

%% file: results.tex
\section{RESULTS}\label{sec:results}

We present several test cases to showcase the accuracy and robustness of our method. All simulations are implemented in Drake and run in a single thread on a system with a Intel Xeon W-11855M processor and 128 GB of RAM. The scene statistics, including the MPM particles count and Eulerian grid spacing $h$, and runtime performance are summarized in Table \ref{table:simulation}. For all simulations, problem \eqref{eq:mpm_optimization} and \eqref{eq:optimization_problem} are solved to convergence with relative tolerance $10^{-6}$. The time step is chosen to be $\Delta t = 0.01$ second unless otherwise specified.

\begin{table}
    \centering
    \fontsize{.8em}{.5em}\selectfont
    \caption{\textbf{Simulation statistics.} Runtime is reported as simulation time in seconds per time step. Since the number of grid nodes with nonzero mass varies as particles move in space, $n_v$ is not constant and we report its average over the entire simulation.}
    \begin{tabular*}{\linewidth}{cccccc}
        \toprule
        \multicolumn{1}{c}{Example} & $h\text{[m]}$ & \makecell{\# of\\particles} &  $n_v$ & \makecell{Runtime} & \makecell{Constraints \\Avg. (Max)}\\  \midrule
        Ball on Slope & 0.2 & 6,777 & 1075.5 & $1.16$ &2.6(6)\\\midrule
        Dough Tearing & 0.02 & 1,647 & 2974.7 & $0.86$ & $113.7(306)$\\\midrule
        Dough Rolling & 0.02 & 5,863 & 5827.3 &$7.65$ &496.3(723)\\\midrule
        Water Pouring & 0.01 & 791 & 1468.1 &0.85 &312.6(606) \\\midrule
        Shake    & 0.01 & 3,456& 1576.0 &1.94 &763.7(1626)\\\midrule
    \end{tabular*}
    \label{table:simulation}
\end{table}

\subsection{Comparison Against Analytic Solutions}
\begin{figure}\centerline{\includegraphics[width=0.8\columnwidth]{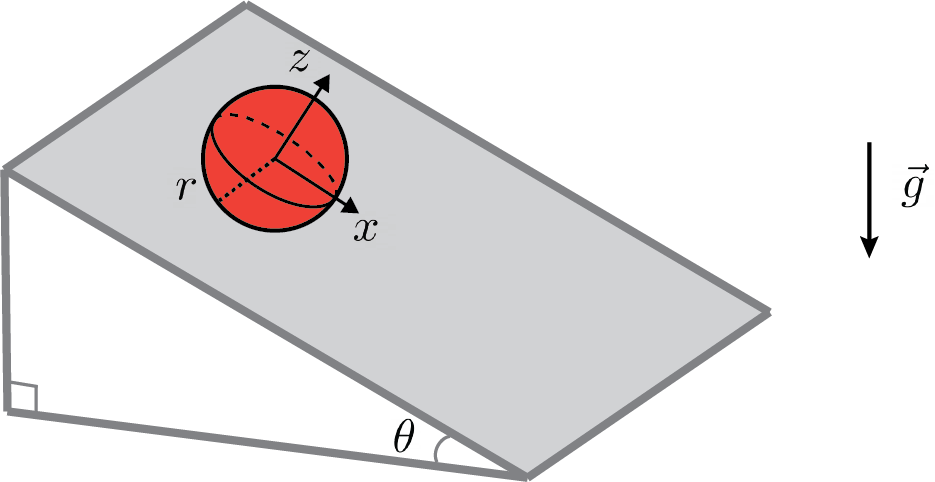}}
\caption{A rigid ball rolling down a rigid slope. The analytical solution in \eqref{eq:roll-sphere-analytical} is the displacement in $x$ direction.}
\label{fig:slope-schematic}
\end{figure}

\begin{figure}
\centerline{\includegraphics[width=1.0\columnwidth]{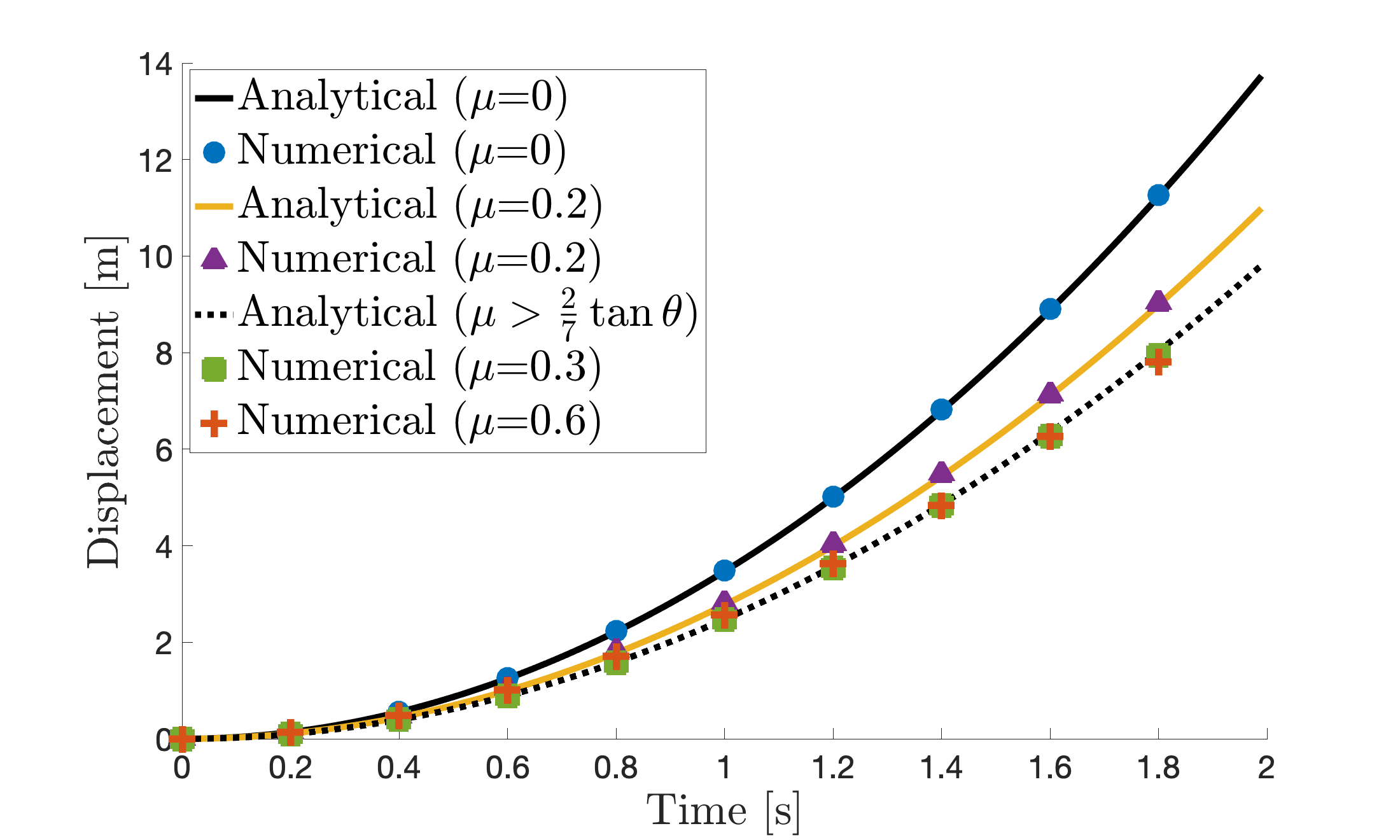}}
\caption{Displacement of the center of mass of a rigid ball along a rigid slope: analytical and numerical solutions. Our results match well with the analytical solution in both the slip mode ($\mu \leq \frac{2}{7}\tan\theta$) and the stick mode ($\mu > \frac{2}{7}\tan\theta$).}
\label{fig:slope-data}
\end{figure}

We first validate our contact model by considering a ball rolling down a rigid slope due to gravity as done in 
\cite{huang2011contact, bib:zhao2023coupled}. The ball is initially tangent to the slope with zero velocity. A schematic illustration of the problem is depicted in Fig. \ref{fig:slope-schematic}. Here we set the angle of the slope to be $\theta = \pi/4$. When both objects are rigid, the problem has an analytical solution \cite{bib:zhao2023coupled}. The tangential displacement of the center of mass of the ball is:
\begin{equation}
     u_x = \begin{cases} 
      \frac{1}{2}gt^2 (\sin\theta - \mu\cos\theta) &  \text{ if } 0\leq \mu \leq \frac{2}{7}\tan \theta, \\
      \frac{5}{14}gt^2 & \text{ if }  \mu > \frac{2}{7}\tan\theta,
   \end{cases}
   \label{eq:roll-sphere-analytical}
\end{equation}
where $\mu$ is the friction coefficient and $g$ is the gravity constant. We simulate the ball as elastic with a high stiffness of $E = 10^8$ $\text{Pa}$ to approximate its rigidity. The Poisson's ratio is set to $\nu=0.2$. The ball has a radius of $0.5$ $\text{m}$ and density $10^3$ $\text{kg/m}^3$. We compare our results with the analytical solutions for $\mu = 0.0, 0.2, 0.3 \text{ and } 0.6.$ As shown in Fig. \ref{fig:slope-data}, our results match well with the analytical solutions.

\subsection{Tearing an Elastoplastic Dough}
MPM is particularly suitable for simulating diverse elastoplastic behavior, including fracture. We simulate a hard cookie dough discretized by MPM particles being teared apart by two KUKA LBR iiwa 7 arms outfitted with custom grippers. The flour dough is modeled with linear corotated model with plasticity, where the Young's modulus is set to $E = 10^5$~Pa, Poisson's ratio is $\nu = 0.4$, yield stress is $\eta = 6\times 10^3$~Pa, and density is $10^3$~$\text{kg/m}^3$ \cite{bib:srikanlaya2017effect}. 
The robots are PD-controlled with a prescribed \textit{grab-lift-tear-lower-release} motion sequence. The resulting states of the dough after the \textit{lift} and \textit{tear} operations are shown in Fig. \ref{fig:dough-tearing}. The complete motion is shown in the supplemental video. Our solver successfully captures the fracture due to permanent plastic deformation.

\begin{figure}
\centerline{\includegraphics[width=1.0\columnwidth]{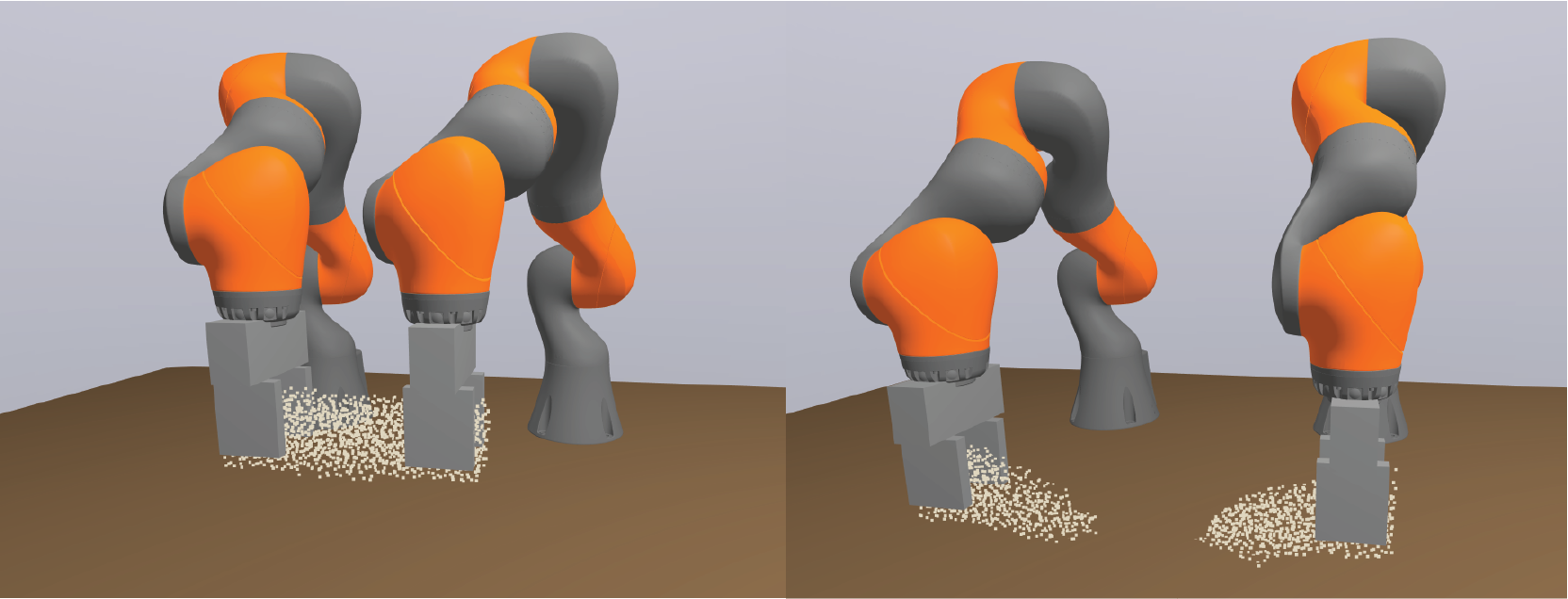}}
\caption{A cookie dough is teared apart into two pieces by robot arms.}
\label{fig:dough-tearing}
\end{figure}

\subsection{Rolling an Elastoplastic Dough}
We demonstrate the two-way coupled frictional contact between rigid and deformable bodies in a challenging dough rolling example (Fig. \ref{fig:dough-rolling}). The soft pizza dough is simulated with Young's modulus $E = 2\times 10^4$~Pa, Poisson's ratio $\nu = 0.4$, yield stress $\eta = 10^3$~Pa, and density is $10^3$~$\text{kg/m}^3$ \cite{bib:srikanlaya2017effect}. 
The rolling pin consists of two co-axial cylinders connected by a revolute joint and is held by two PD-controlled KUKA LBR iiwa 7 arms outfitted with Schunk WSG 50 grippers. The rolling process initiates with the rolling pin pressing vertically into the dough, then moving backward to flatten and extend the dough's shape. Subsequently, the pin is elevated, advanced forward, and slightly rotated before being pressed into the dough once more, executing a forward then a backward roll. In the concluding backward roll, the pin presses deeper into the dough, creating a more distinct imprint. The friction coefficient between the dough and the rolling pin is set to $1.0$ The frictional contact causes the rolling pin to spin and facilitates the rolling action. Readers are referred to the supplemental video for the complete motion sequence.

\subsection{Transferring Liquid}
\begin{figure}
\centerline{\includegraphics[width=1.0\columnwidth]{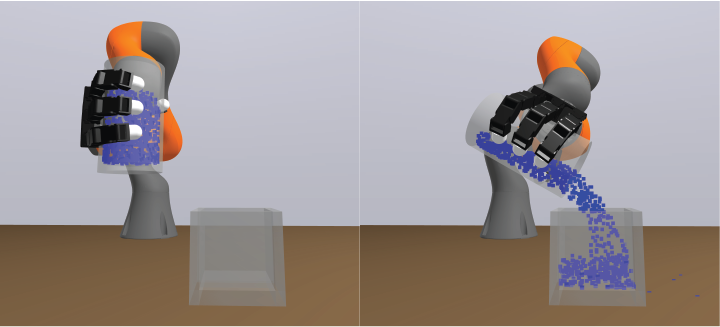}}
\caption{A robot transferring liquid from a mug into a bin.}
\label{fig:liquid}
\end{figure}
Although the quadratic energy density model in \eqref{eq:energy_density} enjoys the property that \eqref{eq:mpm_optimization} can be solved in a single Newton step, our framework is not limited to this model alone and is versatile enough to incorporate a variety of constitutive models for simulating different material behaviors. With a non-convex energy density model, problem \eqref{eq:mpm_optimization} is no longer convex and the our solver might produce a local minimum of $E_\text{MPM}$. However, $\mf{A}_{\text{MPM}}$ is still SPD and consequently problem \eqref{eq:optimization_problem} is still convex and admits a unique global minimum. In Fig. \ref{fig:liquid} we show a simulation of a robot transferring liquid simulated with the Equation-of-State constitutive model from a mug to a bin \cite{bib:monaghan1994simulating, bib:tampubolon2017multi}. The robot is modelled as a KUKA LBR iiwa 7 arm equipped with an anthropomorphic Allegro hand. The material parameters of the liquid are set as in \cite{bib:tampubolon2017multi}. 

\subsection{Comparison with ManiSkill2}
We compare our solver with ManiSkill2 \cite{bib:gu2023maniskill2}, the state-of-the-art embodied AI environment that supports two-way coupled rigid body MPM simulation. 
In ManiSkill2, frictional contact between MPM and rigid bodies are resolved through explicit impulse exchange at each time step. The dynamics of MPM is also integrated explicitly by taking sub-steps within the rigid body time steps. Such explicit treatment simplifies implementation and is easily amenable to extensive parallelization on GPUs. Despite these advantages, the explicit approach can be inaccurate in handling friction and is prone to instabilities in contact intensive scenarios \cite{bib:erleben2007velocity, bib:li2020ipc}. 

\begin{figure}
\centering
\begin{subfigure}{.38\columnwidth}
  \centering
  \raisebox{2mm}{
  \includegraphics[width=0.96\columnwidth]{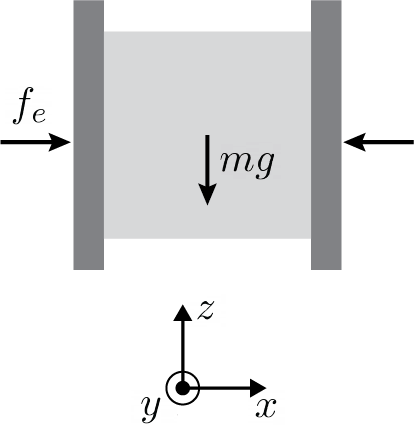}}
  \caption{}
  \label{fig:box-schematic}
\end{subfigure}%
\begin{subfigure}{.62\columnwidth}
  \centering
  \includegraphics[width=0.96\columnwidth]{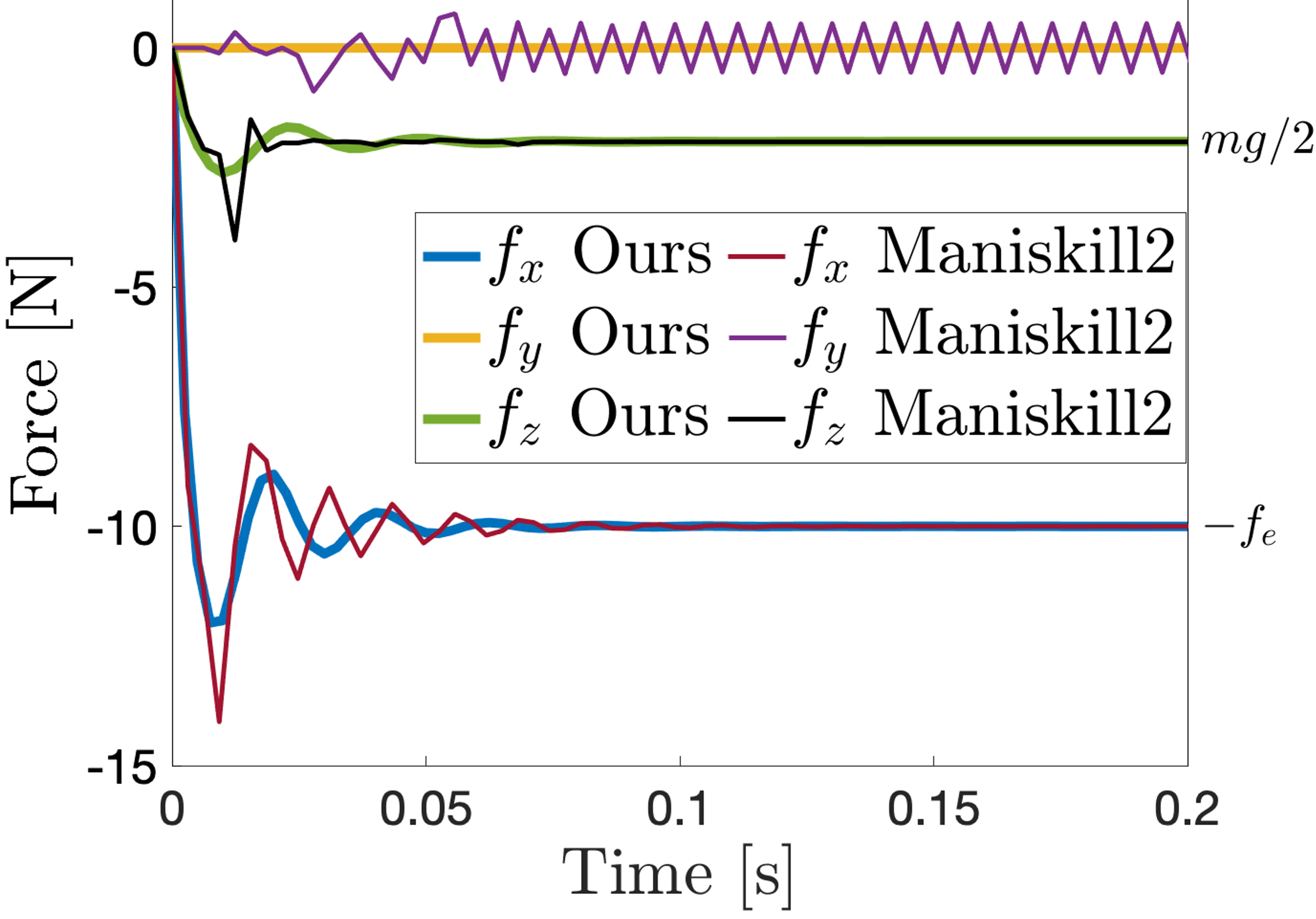}
  \caption{}
  \label{fig:single-box-compare}
\end{subfigure}
\caption{(a) A normal force of $f_e = 10$~N is exerted in the $x$-direction on each of the rigid panels. (b) Contact forces applied by the elastic cube on the left rigid panel. The corresponding analytical values at steady state are marked on the right.}
\label{fig:test}
\end{figure}

To illustrate this, we simulate an elastic cube with side length $0.1$~m, mass $m=0.4$~kg, Young's modulus $E = 10^5$~Pa and Poisson's ratio $\nu = 0.4$ compressed by two rigid panels exerting a constant force of $10$~N each (see Fig.~\ref{fig:box-schematic}). The friction coefficient is set to $\mu=0.8$ so that the friction forces are sufficient to counteract the cube's weight. We simulate this scenario with both our method and the method proposed by ManiSkill2 with $\Delta t = 10^{-3}$~second, with 25 substeps for ManiSkill2's explicit MPM integration. The contact forces exerted by the cube on the left panel are depicted in Fig.~\ref{fig:single-box-compare}. The normal and $z$-direction friction forces converge to the analytical solutions after the initial transient period for both methods. However, ManiSkill2's approach exhibits high-frequency oscillations in the $y$-direction friction force as the explicit method struggles to achieve the equilibrium in stiction. The oscillation persists with the same magnitude even as we decrease $\Delta t$ to as small as $2.5 \times 10^{-5}$~second.

\begin{figure}
\centerline{\includegraphics[width=1.0\columnwidth]{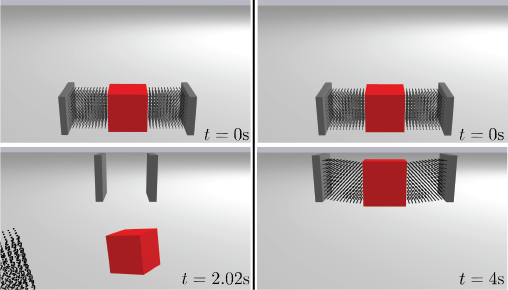}}
\caption{Two actuated rigid panels lift and shake rigid and elastic cubes. Our method robustly completes the task (lower right), whereas the explicit method fails even with small time steps (lower left).} 
\label{fig:shake}
\end{figure}

This inaccuracy is magnified in more complex tasks. As shown in Fig.~\ref{fig:shake}, we prescribe a \textit{close-lift-shake} motion sequence for the rigid panels to pick up a red rigid cube positioned between two elastic cubes. To make the problem more challenging, we set the density of the rigid cube to be $150$ times greater than that of the elastic cubes. Our method robustly completes the grasp with $\Delta t = 0.01$~second when sufficient normal forces are applied on the panels. In contrast, the method proposed by ManiSkill2 fails the grasping task due to instabilities in the contact forces, a problem that persists even when $\Delta t$ is reduced to $2.5 \times 10^{-5}$~second. We refer readers to the supplemental video for the comparison of dynamics.

%% file: future_work.tex
\section{Limitations and Future Work}

\textbf{Runtime performance:} 
In this work, we adopt a serial implementation of our method, prioritizing accuracy and algorithm design. We note that some of the most time-consuming routines such as solving for problem \eqref{eq:mpm_optimization} and \eqref{eq:optimization_problem} can greatly benefit from a parallel implementation where matrix-free methods are utilized \cite{bib:huang2023gipc}. However, the impact of such an approach on convergence remains an open question. Exploiting parallel computing on contemporary hardware is the current focus of our ongoing research.

\textbf{Discrete contact detection:} Our method employs discrete contact detection at each time step, which raises the possibility that a high-speed particle might pass through a thin rigid body within a single time step without being detected. However, this concern is somewhat alleviated by the observation that objects typically do not move at high speeds in robotic manipulation tasks.

\textbf{Rotational invariance:} The linear corotated elastoplastic model we proposed shares the same stress-strain relationship with the constitutive model from \cite{bib:han2023convex} within the elastic limit. Therefore, it is not rotationally invariant and suffers from similar artifacts in scenarios with fast rotational motion discussed in \cite{bib:han2023convex}.


%% file: conclusion.tex
\section{Conclusion}

We introduced a novel convex formulation that seamlessly integrates MPM with rigid bodies via frictional contact. Additionally, we developed a new elastoplastic constitutive model that enhances the efficiency of solving the convex formulation. To demonstrate the effectiveness of our method, we presented validation results and comparison against leading alternative solvers. We identified limitations of our approach and outlined potential directions for further research. Finally, we incorporate our method into the open source robotics toolkit Drake \cite{bib:drake}, and hope that the simulation and robotics communities can benefit from our contribution.